\documentclass[conference]{IEEEtran}
\IEEEoverridecommandlockouts
\usepackage{cite}
\usepackage{amsmath,amssymb,amsfonts}
\usepackage{algorithmic}
\usepackage{graphicx}
\usepackage{textcomp}
\usepackage{xcolor}
\usepackage{gensymb}
\usepackage[hyphens]{url}

\def\BibTeX{{\rm B\kern-.05em{\sc i\kern-.025em b}\kern-.08em
    T\kern-.1667em\lower.7ex\hbox{E}\kern-.125emX}}
\usepackage{tabu}
\begin{document}

\title{Perspective-consistent multifocus multiview 3D reconstruction of small objects\\
}

\author{\IEEEauthorblockN{Hengjia Li}
\IEEEauthorblockA{\textit{College of Engineering and Computer Science} \\
\textit{Australian National University}\\
Canberra, Australia \\
hengjiali0625@gmail.com}
\and
\IEEEauthorblockN{Chuong Nguyen*}
\IEEEauthorblockA{\textit{Imaging and Computer Vision Group} \\
\textit{CSIRO DATA61}\\
Canberra, Australia \\
chuong.nguyen@csiro.au}}



\maketitle

\begin{abstract}
Image-based 3D reconstruction or 3D photogrammetry of small-scale objects including insects and biological specimens is challenging due to the use of high magnification lens with inherent limited depth of field, and the object's fine structures and complex surface properties. Due to these challenges, traditional 3D reconstruction techniques cannot be applied without suitable image pre-processings. One such preprocessing technique is multifocus stacking that combines a set of partially focused images captured from the same viewing angle to create a single in-focus image. Traditional multifocus image capture uses a camera on a macro rail. Furthermore, the scale and shift are not properly considered by multifocus stacking techniques. As a consequence, the resulting in-focus images contain artifacts that violate perspective image formation. A 3D reconstruction using such images will fail to produce an accurate 3D model of the object. This paper shows how this problem can be solved effectively by a new multifocus stacking procedure which includes a new Fixed-Lens Multifocus Capture and camera calibration for image scale and shift. Initial experimental results are presented to confirm our expectation and show that the camera poses of fixed-lens images are at least 3-times less noisy than those of conventional moving lens images. 
\end{abstract}


\section{Introduction}
3D reconstruction of small objects including insects and biological specimens is challenging due to the use of high magnification lens with limited depth of field, fine features and complex surface properties. Recent advancements \cite{nguyen2014capturing, strobel2018automated} show that photogrammetry or image-based multiview 3D reconstruction can be applied with some success to create true-color 3D models of small specimens. Solutions enabling images of small specimens to be reconstructed in 3D include: a two-axis turntable combined with macro rail and macro photography to capture multifocus multiview images \cite{nguyen2014capturing}, calibration target \cite{nguyen2014capturing, strobel2018automated}, multifocus image stacking \cite{Wiki-Focus-stacking}, scale-shift calibration and automatic background masking \cite{strobel2018automated}, and modern photogrammetry software \cite{seitz2006comparison, 3DSOM, Agisoft-Photoscan}. Despite multiple techniques to tackle different issues of image-based 3D reconstruction of small specimens, obtaining an accurate 3D model is still difficult. Particularly, camera pose estimation from stacked in-focus images is still unreliable, and it causes poor quality 3D reconstruction. To better estimate camera poses, a special calibration object, such as a textured sphere, needs to be used \cite{strobel2018automated} and camera poses are precomputed. However, this workaround solution restricts to predetermined camera positions and relies on the high repeatability of motorized hardware.

In this paper, we identify multifocus image stacking as one major source of error causing unreliable 3D reconstruction. Multifocus stacking combines partially in-focus images by automatically detecting in-focus regions from the images and merging them to generate an in-focus image. The resulting image is just a good-looking mosaic image which does not accurately represent a perspective projection. Scale-shift calibration proposed by \cite{strobel2018automated} partially corrects this problem by scaling and shifting a set of partially in-focus images before being blended into a stacked in-focus image. However, as the original image set are captured at different camera lens positions, the blended in-focus image never accurately represents a perspective projection. This is illustrated by the bottom of Figure \ref{fig:perspective_distortion}. The use of such stacked in-focus images breaks the major assumption in 3D reconstruction algorithms \cite{seitz2006comparison} that each input image representing a single perspective pinhole projection at a single camera position and orientation (called camera pose). As a result, a 3D multiview reconstruction using such in-focus images from different viewing angles fails to generate an accurate 3D model of the object. 

This paper proposes an effective solution to capture multifocus images, and enable multifocus stacking to produce an in-focus image representing a single perspective projection for accurate 3D reconstruction of small objects. Our contributions include: a) a newly proposed technique called Fixed-Lens Multifocus Capture to capture suitable multifocus images, b) a new multifocus stacking procedure to create an in-focus image that correctly represents a perspective projection. An overview of the concept of pinhole camera model, the problem of perspective inconsistency, depth of focus and multifocus stacking is introduced in section II. Section III describes our proposed method of multifocus multiview 3D reconstruction: the fixed lens setup, camera calibration, multifocus stacking based on Guided Filtering, and background masking. Section IV presents experimental results showing the effectiveness of the proposed method versus the conventional method. Finally, section V concludes the paper with our findings. 

\section{Perspective image formation and depth of focus effect}

\subsection{Pinhole camera and perspective image formation}

\begin{figure}[htbp]
\centerline{\includegraphics[width=0.8\columnwidth]{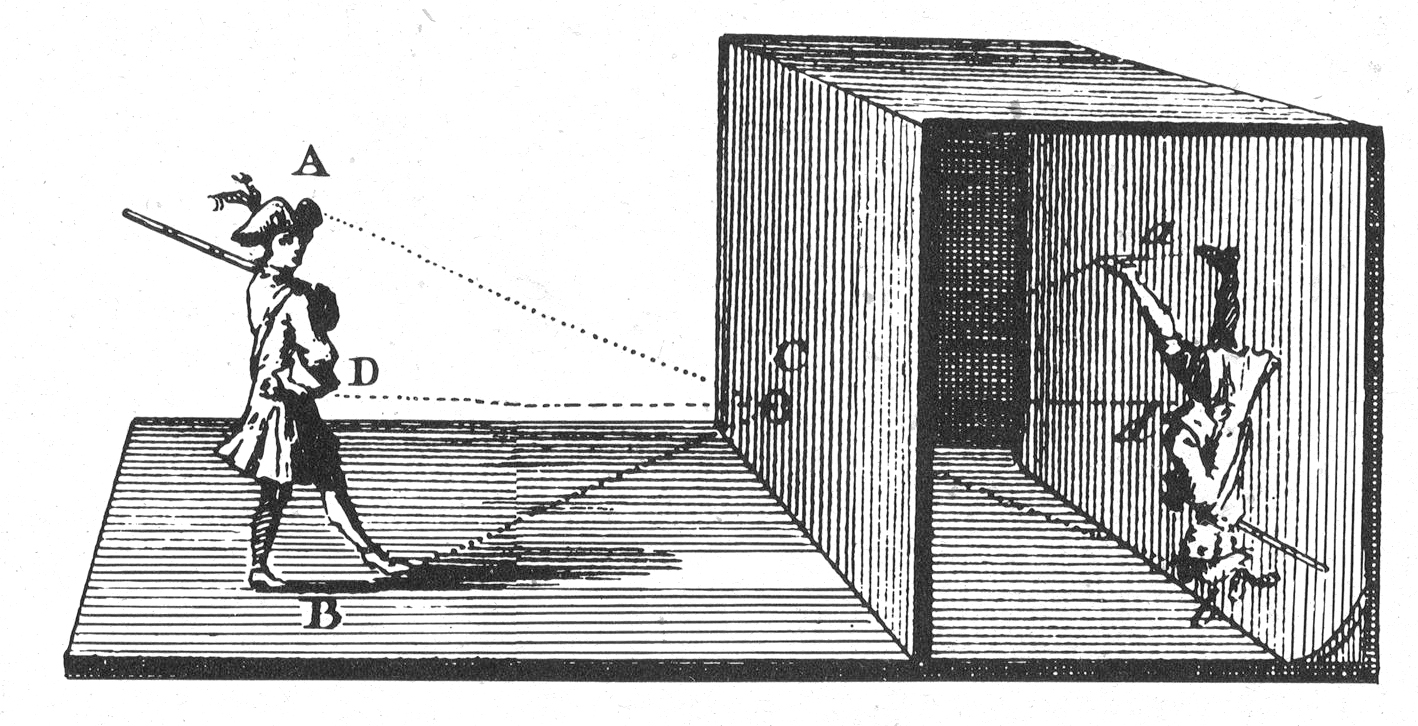}}
\centerline{\includegraphics[width=0.8\columnwidth]{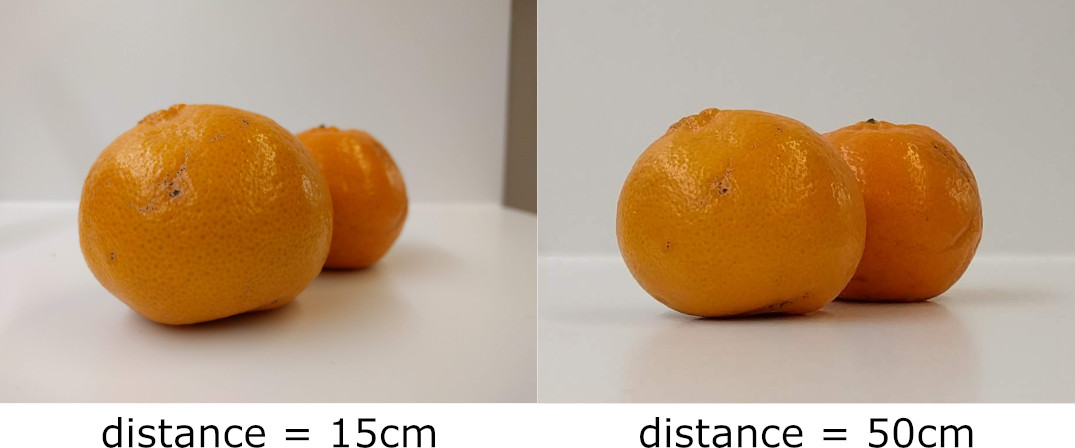}}
\caption{Top: Pinhole camera principle and perspective projection\cite{Wiki-Camera_obscura} where all straight light rays go through this hole (image in public domain). Pinhole represents the camera centre for image formation. Bottom: the pair of oranges captured by same camera of 5mm focal length at different distances leading to different perspective image formations (or distortions). }
\label{fig:perspective_distortion}
\end{figure}

The firstly invented camera is the pinhole camera as shown in the top of Figure \ref{fig:perspective_distortion} where the pinhole C is the centre of image formation allowing light rays passing straight through. Modern cameras use lenses instead of the pinhole to improve image quality, but it adds the depth of focus where a film or image sensor is located to capture a clear in-focus image. Due to the image capture process of a camera, a transformation between the 3D world to 2D image happens where the depth is missing in the captured image. Furthermore, parts in the scene appear at different scales or magnifications depending on their distance to the camera lens. This is demonstrated by the bottom of Figure \ref{fig:perspective_distortion} where the images of the same object are captured at two different distances. As the camera lens moves, the centre C of image formation moves. This leads to the changes in the relative distances (and magnification) between parts of the scene to camera centre. As a result, the change of relative scales of different parts of the scene causes different perspective distortions.

\subsection{Depth of focus, depth of field, and multifocus stacking}
\begin{figure}[htbp]
\centerline{\includegraphics[width=0.9\columnwidth]{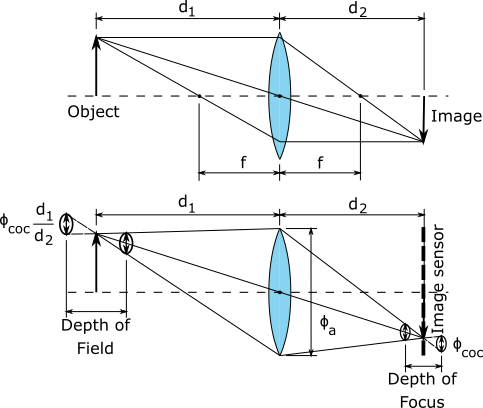}}
\caption{Top: image formation through a lens of focal length $f$. After passing the lens, rays parallel to optical axis go through focal point and rays going through focal point becomes parallel to optical axis. Bottom: image sensor is used to capture an object image at a position within the depth of focus where a point source grows into a circle of confusion \cite{Wiki-Circle_of_confusion} of diameter $\phi_{coc}$. A captured image is considered as out of focus if image sensor is placed outside the depth of focus or the object is located outside of the depth of field.}
\label{fig:depth_of_focus}
\end{figure}

Due to the use of lenses in modern cameras, image quality is significantly better, but the depth of focus is introduced in the image. As shown in the top of Figure \ref{fig:depth_of_focus}, an image of the object at distance $d_1$ forms a image at distance $d_2 = \frac{d_1 f}{d_1 - f}$, where $f$ is the lens focal length. If an image sensor is placed at this distance (i.e. at back focal plane), the image will be clear and in focus. 

In reality, an image is considered in focus if the image of a point remains smaller than a circular dot called the circle of confusion (COC) with an empirical diameter $\phi_{coc}$. From \cite{Wiki-Circle_of_confusion}, $\phi_{coc}$ is chosen approximately 0.1\% of the mean of the width and height of the image sensor. For example, for a 35mm sensor format, $\phi_{coc}$ is chosen to be 0.025mm. Strictly speaking, $\phi_{coc}$ should be selected as the larger value of the size of an image pixel and optical resolution of the lens. 

The distance range where the image remains in focus is called the depth of focus $DOFocus$. Using the proportional relationship, one can obtain the depth of focus as:

\begin{align}
    \frac{DOFocus}{2 d_2} &= \frac{\phi_{coc}}{\phi_{a}} \\
    \Rightarrow DOFocus &= 2\frac{\phi_{coc}}{\phi_{a}}d_2 \\
    \text{or } DOFocus &= 2 \frac{f_{number} \phi_{coc}}{f}d_2
\end{align}
\noindent where $\phi_{a}$ is the diameter of the lens aperture, and $f_{number} = \frac{f}{\phi_{a}}$. 

This depth of focus translates to the depth of field $DOField$ where objects stay within so that their image is in-focus. Assuming the circle of confusion $\phi_{coc}$ is back projected into the scene to the size $\frac{\phi_{coc}}{M}$ where $M=\frac{d_2}{d_1}$ is lens magnification, the proportional relationship gives:
\begin{align}
    \frac{DOField}{2 d_1} &= \frac{\phi_{coc}}{M \phi_{a}} \\
    \Rightarrow DOField &= 2 \frac{\phi_{coc}}{M \phi_{a}}d_1 \\
    \text{or } DOField &= 2 \frac{d_1^2\phi_{coc}}{d_2 \phi_{a}} = 2 \frac{d_1 f \phi_{coc}}{(d_1 - f) \phi_{a}}
\end{align}

To capture scenes at different distances, the position of the lens and/or the image sensor needs to be adjusted to put the image into focus. 


\begin{figure}[htbp]
\centerline{\includegraphics[width=0.7\columnwidth]{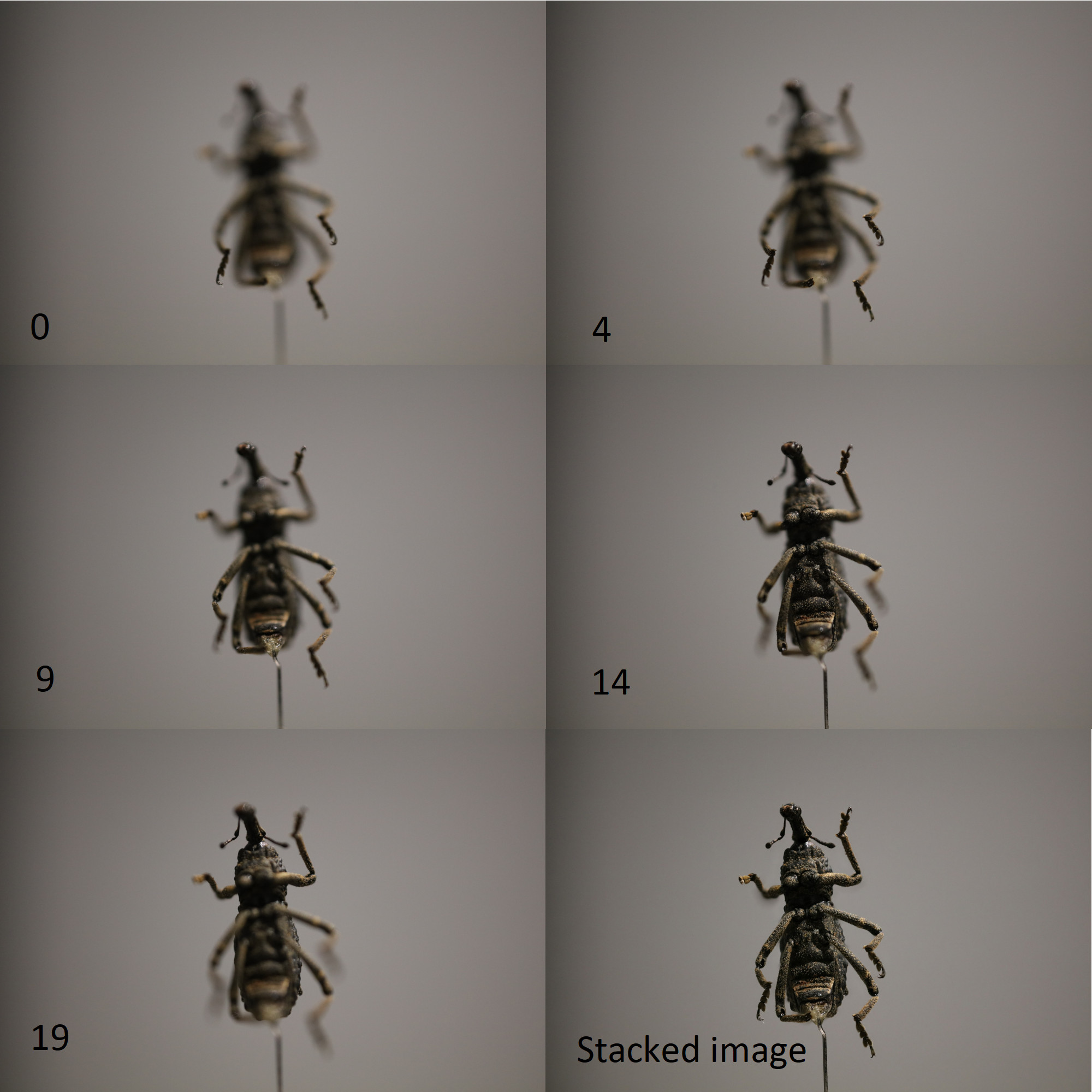}}
\caption{A set of 20 partially out-of-focus images (only 5 of them shown here) captured at different camera positions are stacked to produce an in-focus image. The stacked image however does not represent a single perspective image formation due to moving lens.}
\label{fig:multifocus}
\end{figure}

For an object with parts at different distances, only some parts are in focus and the other parts are out-of-focus. This is especially prevalent in macro imaging where the size of the scene is tens of centimeter or smaller and high magnification lens is used.
Figure \ref{fig:multifocus} shows images captured as a camera (and lens) moves towards the specimen at constant incremental depth at the same viewing angle. Multifocus image stacking produced an in-focus image representing that single viewing angle.

Conventional multifocus capture is to move camera and lens together. The step size for macro rail is equivalent to the depth of field. However, the recommended macro rail step size is 50\% of the depth of field in this case to allow 50\% overlapping between in-focus regions of successive images to help guide image registration for optimisation-based multifocus stacking algorithms.

By using such multifocus stacked images captured at different angles, 3D reconstruction of small objects becomes possible as reported by \cite{nguyen2014capturing, strobel2018automated}. However, such in-focus stacked images do not correctly represent perspective image formation, leading to reconstruction errors or failures. This becomes more severe when the size of the objects becomes smaller, i.e. a few millimeters long or smaller.

\section{Multifocus image stacking with fixed perspective image formation}

\subsection{Fixed-Lens Multifocus Capture}

\begin{figure}[htbp]
\centerline{\includegraphics[width=0.8\columnwidth]{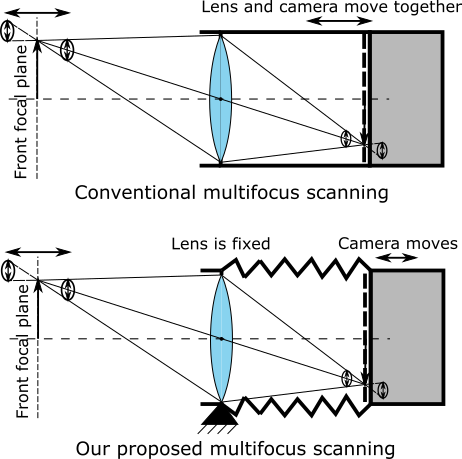}}
\caption{Conventional multifocus capture (top) and proposed Fixed-Lens Multifocus Capture (bottom) for multifocus image capture. For the conventional scanning, the movement of the camera and lens is the same as that of the front focal plane. For the proposed scanning, the movement of the camera is smaller than that of front focal plane if the lens magnification is smaller than 1, and vice versa.}
\label{fig:multifocus_scannings}
\end{figure}

Conventional camera setup captures multifocus images by moving the camera and lens together as shown in the top of Figure \ref{fig:multifocus_scannings}. As the whole camera moves, the camera centre moves, and perspective view moves with it. To avoid moving the perspective view of the camera, we proposed that the lens is fixed and only the image sensor moves during scanning as shown in the bottom of Figure \ref{fig:multifocus_scannings}. 
In this case, the centre of perspective image formation does not change. Although the image size of the whole scene becomes larger when the film is moving away from the lens, the relative scale or magnification of different parts of the scene stays the same.

\begin{figure}[htbp]
\centerline{\includegraphics[width=\columnwidth]{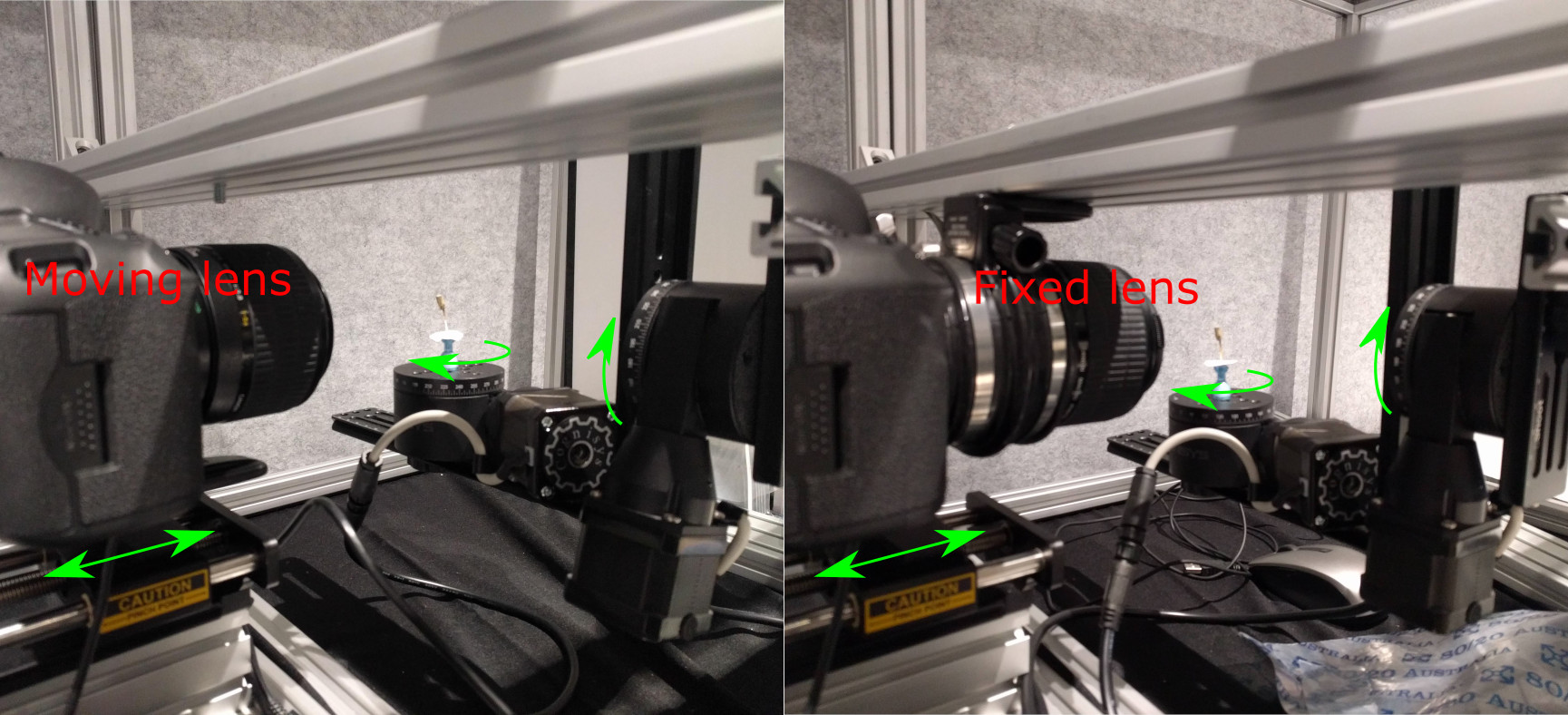}}
\caption{Lens and pan-tilt setup for multifocus multiview image capture of small objects. The camera is mounted on a macro rail fixed to the lower frame for multifocus capture. A macro lens is attached to a camera for conventional moving lens setup (left) or to a stationary frame for our proposed fixed lens setup (right). For the fixed lens setup, an expandable rubber duct connects the macro lens and the camera body. The object/specimen is pinned to a fridge magnet on a pan-tilt turntable for multiview capture.
}
\label{fig:new_scanner_setup}
\end{figure}

An example of camera lens setup for the proposed Fixed-Lens Multifocus Capture is shown on the right of Figure \ref{fig:new_scanner_setup} as compared to the conventional one (left). The lens is fixed to the upper frame while the camera body is moved by a macro rail to capture multiple partially focused images. A rubber duct connects the lens and the camera body to prevent ambient light and dusts entering the camera sensor chamber. A small flower is mounted on a 2-axis turntable to capture images at different pan-tilt angles.

\subsection{Camera calibration to measure image scale and shift}

Images captured at different camera distances from the lens have different magnifications or scales. Furthermore, the direction of the camera's movement is not generally aligned with the lens optical axis, causing a relative shift in the image.
To account for such changes, a camera calibration is performed to measure the relative scale change and shift from one image to another. This can be carried out by capturing multifocus images of a known calibration target placed perpendicular to the optical axis of the lens. The relative scale and shift can be measured from the relative positions of the control points (circles) between the images. Once the relative scales of the scene in partially focused images are determined, image mapping is applied to remove the scale change and shift between images captured from the same viewing angle.

\begin{figure}[htbp]
\centerline{\includegraphics[width=0.8\columnwidth]{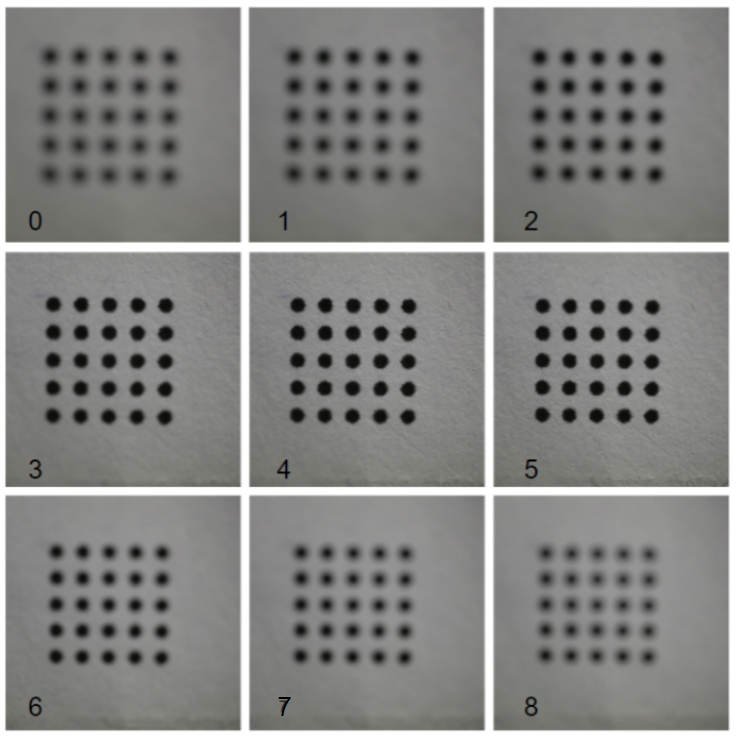}}
\caption{A set of 9 target images (from 0 to 8) captured at different camera positions for camera calibration. The camera moves toward the calibration target and its focal plane passes through the calibration target. Thus, the captured images are from out-of-focus (blurry) to in-focus (sharp) and become out-of-focus (blurry) again. The scale of the grids gets smaller as the cameras comes toward the target. There is also a small shift from top left to bottom right of the image. }
\label{fig:dot_pattern}
\end{figure}

Figure \ref{fig:dot_pattern} shows calibration images captured as a camera moves perpendicularly towards the calibration target at constant incremental depth at the same viewing angle, and this camera motion will be repeated during the scan of a specimen. The image of calibration pattern represents the scale and location of the object image as the camera moves to capture multifocus images. A small in-plane shift between images can be observed in additional to the change of scale.

\begin{figure}[htbp]
\centerline{\includegraphics[width=0.7\columnwidth]{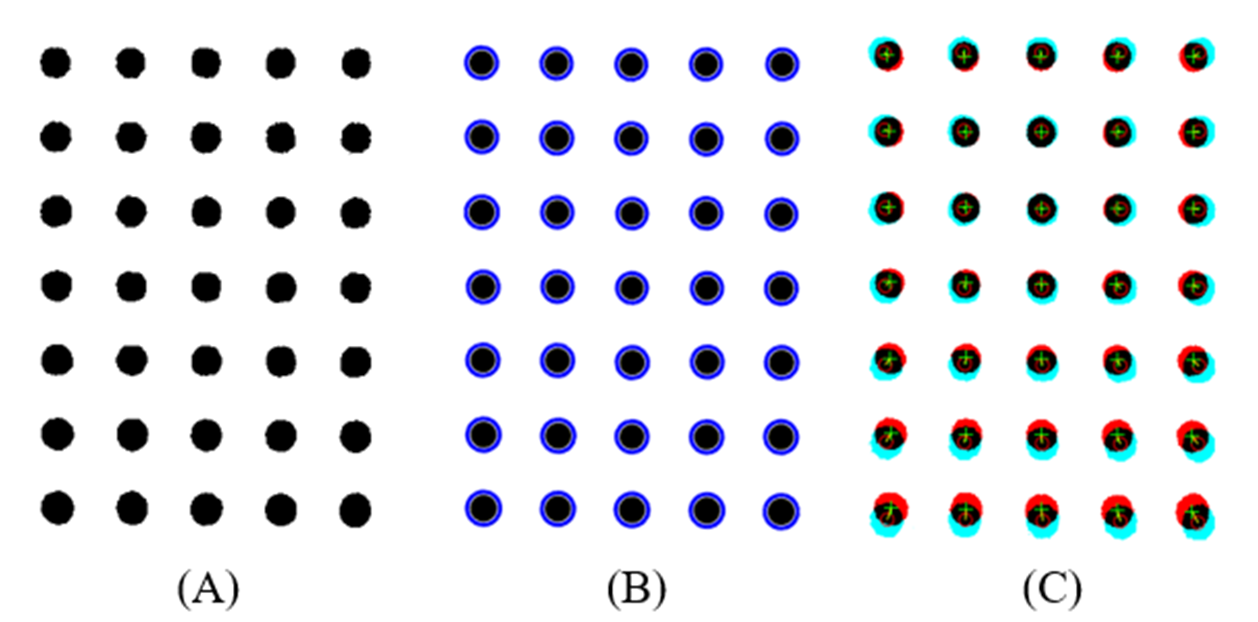}} 
\caption{Calibration process for image stacking. (A) image of calibration target with black circular dots. (B) detected dots shown as blue (C) Overlay of circular dots detected from an image (red) and those from a reference image (green). A homography matrix is computed from the corresponding points and used to transform the images captured at a similar camera depth position of red dots to align with the reference image. }
\label{fig:circle_detection}
\end{figure}

Instead of measuring directly the scale and the in-plane shift between an image and a reference image as in \cite{strobel2018automated}, we compute a homography matrix from the points correspondingly and use this to transform an image to align with the reference image before multifocus stacking. Figure \ref{fig:circle_detection} shows the calibration dots (A) are detected as in (B) and compared with a set from its reference image (C). A homography is computed from point corresponding between that image and the reference image (usually the one at the middle of the set). The homography matrix $H$ is defined as a linear transformation between the positions $X_{i}$ of the detected calibration dots on the i-th image and those $X_{r}$ on the reference image:
\begin{align}
X_{r} &= HX_{i}  \\
\begin{bmatrix}
 x_{r}
\\ y_{r}
\\ w_{r}
\end{bmatrix}  
&=
\begin{bmatrix}
 h_{1}^TX_{i}
\\ h_{2}^TX_{i}
\\ h_{3}^TX_{i}
\end{bmatrix}
\end{align}

Rewrite this as:

\begin{align}
X_{r} \times HX_{i}=\begin{bmatrix}
 y_{r}h_{3}^TX_{i}-w_{r}h_{2}^TX_{i}
\\ w_{r}h_{1}^TX_{i}-x_{r}h_{3}^TX_{i}
\\ x_{r}h_{2}^TX_{i}-y_{r}h_{1}^TX_{i}
\end{bmatrix}=0
\end{align}

Factorise this to:

\begin{align}
\begin{bmatrix}
0 & -w_{r}X_{i}^T & y_{r}X_{i}^T\\ 
w_{r}X_{i}^T & 0 & -x_{r}X_{i}^T\\ 
-y_{r}X_{i}^T & x_{r}X_{i}^T & 0
\end{bmatrix}
\begin{bmatrix}
h_{1}\\ 
h_{2}\\ 
h_{3}
\end{bmatrix}=0
\label{eq:1}
\end{align}

The homography matrix can be solved by singular value decomposition of Equation \ref{eq:1}.

After the captured images from the same viewing angle are transformed to align to the reference image using the homography obtained from corresponding calibration images, these transformed images can be stacked or blended to create an in-focus image.

\subsection{Multifocus image stacking of transformed images}

\begin{figure}[htbp]
\centerline{\includegraphics[width=\columnwidth]{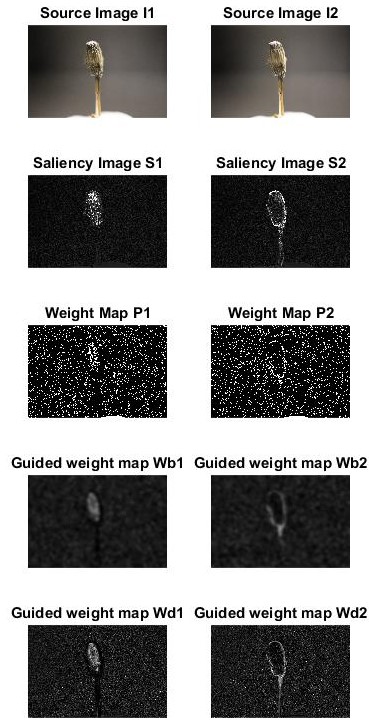}}
\caption{ Example outputs of weight-map construction from a pair of input images that have different regions in focus. From top to bottom: a) The source images I1 and I2 with different in-focus regions. b) The saliency maps S1 and S2 highlight image regions in focus. c) The weight maps P1 and P2 from the comparison between the saliency maps. d) Guided weight maps Wb1 and Wb2 for blending the base layer. e) Guided weight maps Wd1 and Wd2 for blending the detail layer.}
\label{fig:weight maps}
\end{figure}

To stack the multifocus images, we apply a guided-filter based blending algorithm focusing on local sharpness/saliency of images \cite{kaiming2013guidedfilter, Shuntao2013}. The local sharpness can be computed by convolving the source image with a Laplacian of Gaussian filter. As shown in Figure \ref{fig:weight maps}, the saliency values in images S1 and S2 reveal the in-focus regions in source images I1 and I2. Therefore, the in-focus regions in each image can be selected according to the saliency values and a linear combination of the detected in-focus regions leads to the all-in-focus images as blending results.

The linear combination of in-focus regions requires weight maps to blend images in particular proportions. For each pixel in the stack images, its corresponding weight is 1 if the pixel has the highest sharpness value among all stack images, otherwise its weight is zero. Thus, the weight map can indicate the index of image that has an in-focus point among all pixels, and it can be written as:

\begin{align}
P_{n}^{k} = \left\{\begin{matrix} 1, if S_{n}^{k} = max(S_{1}^{k},S_{2}^{k},......,S_{N}^{k})
\\ 0, otherwise
\end{matrix}\right.
\label{eq:weight_map}
\end{align}
where ${N}$ is the number of source images and $S_{n}^{k}$ represents the saliency value of the pixel $K$ in the $n$-th image. Figure \ref{fig:weight maps} shows example weight maps (P1 and P2) computed from a pair of images having the different region in focus. The generated weight maps are  noisy at this stage and they can be cleaned by guided filtering.

Guided filtering \cite{kaiming2013guidedfilter} has a strong edge-preserving property. It enhances the edges in an input image according to another guidance image. In theory, the guided filter is defined as linear translation-variant. The filtering output $F$ is a linear transformation of the guidance image $I$ in a window $w_{k}$ of size $(2r+1)(2r+1)$ and centred at the pixel $k$:
\begin{align}
    \ F_{i} = a_{k}I_{i}+b_{k}, \forall i \in w_{k}
\end{align}
where ($a_{k},b_{k}$) are linear coefficients and $i$ denotes a pixel in window $w_{k}$. The linear coefficients are determined by minimizing the difference between the output image $F$ and the input image $P$. Therefore, a cost function can be written as: 
\begin{align}
    \ E(a_{k},b_{k}) =\Sigma_{\forall i \in w_{k}}((a_{k}I_{i}+b_{k}-P_{i})^2+\varepsilon a_{k}^2
\end{align}
where $\varepsilon$ denotes a regularization parameter preventing $a_{k}$ from being too large. By linear regression, the solution of the above equation can be obtained as:
\begin{align}
    \ a_{k}= \frac{\frac{1}{|w|}\Sigma_{i \in w_{k}}I_{i}P_{i}-\mu_{k}\overline{P_{k}}}{\sigma_{k}^2+\varepsilon}\\
    \ b_{k}= \overline{P_{k}}-a_{k}\mu_{k}
\end{align}
Here $\mu_{k}$ and $\sigma_{k}$ are the mean and variance of I in $w_{k}$ correspondingly. $|w|$ denotes the number of pixels in the window $w_{k}$ and $\overline{P_{k}}$ is the mean value of $F_{i}$  in the window.  Therefore, the filtering output can be solved. However, there are multiple windows $w_{k}$ that contains pixel i, so the value of filtering output is not constant while computing for different windows. The strategy applied here is to average all possible values of the output and the output of guided filtering becomes:
\begin{align}
    \ F_{i} = \frac{1}{|w|} \Sigma_{k:i \in w_{k}} (a_{k}I_{i}+b_{k}) = \overline{a_{i}}I_{i}+\overline{b_{i}}\\
    \overline{a_{i}} = \frac{1}{|w|} \Sigma_{k\in w_{k}} a_{k}\\
    \overline{b_{i}} = \frac{1}{|w|} \Sigma_{k\in w_{k}} b_{k}
\end{align}

To use guided filtering to refine the weight maps, we apply a two-scale image decomposition as defined in \cite{Shuntao2013}. Each source image is decomposed into a base and a detail layer. The base layer is obtained by average filtering on the source image, and it extracts large-scale variations in image intensity. The base layer can be computed as a convolution:
\begin{align}
B_{n}= I_{n}*Z
\end{align}
where $I_{n}$ denotes the $n$-th source image and $Z$ is the average filter. 
The detail layer is obtained by subtracting the base layer from the source image, and thus it keeps the small-scale texture information. The detail layer can be computed as:
\begin{align}
D_{n}= I_{n}-B_{n}
\end{align}
Then, the obtained weight maps $P_{n}$ (equation \ref{eq:weight_map}) are fed into the guided filter as input, while the source images $I_{n}$ are applied as guidance. Therefore, the guided weight maps ${W_{n}}^{B}$ and ${W_{n}}^{D}$ can be generated for the base and dense layers:
\begin{align}
{W_{n}}^{B} = G_{r_{1},\varepsilon _{1}}(P_{n},I_{n})\\
{W_{n}}^{D} = G_{r_{2},\varepsilon _{2}}(P_{n},I_{n})
\end{align}

\noindent where $r_{1},r_{2},\varepsilon _{1}$ and $\varepsilon _{2}$ are parameters of the guided filter. Meanwhile, the values of the $N$ guided weight maps are normalized such that they sum to one at each pixel $K$. These normalized weight maps preserve clearer edges of sharp regions from the source images comparing to the original weight maps,  see Wb1, Wb2 Wd1 and Wd2 in Figure \ref{fig:weight maps} for examples. Then, the base and detail layers of different source images can be blended together according to the weight maps:
 \begin{align}
 \bar{B} = \sum_{n = 1}^{N}W_{n}^{B}B_{n}\\
\bar{D} = \sum_{n = 1}^{N}W_{n}^{D}D_{n}
 \end{align}
 In the end, the fusing result  $O$ can be generated by combining the blended base layer $\bar{B}$ and the blended detail layer $\bar{D}$:
\begin{align}
O = \bar{B}+\bar{D}.
\end{align}

Figure \ref{fig:weight maps} shows an example weight map construction process for two source images. However, it usually needs multiple multifocus images for recovering a successful blending result. In that case, the guided-filter based blending algorithm still works in the same manner, and the only difference is that it takes more iterations to complete fusion. 

\subsection{Background masking for multiview 3D reconstruction}

Multiview 3D reconstruction optionally accepts masks or object silhouettes to ignore image region belong to the background to speed up the process and to improve the reconstruction accuracy. Automatic background segmentation can lead to errors and often requires manual input to correct them. To improve automatic background segmentation, \cite{strobel2018automated} proposed to use two images, one in normal front lighting and one with a strong back light. The back light is a uniform light source such as a light box to produce a clear contrast between background area and object area. As a result, automatic segmentation of image with strong back light is very accurate and efficient without the need for manual correction. The only drawback of capturing separate images with back light is the doubling of the data storage and the time to capture and preprocess images.

\section{Experiments and discussion}

\subsection{Experiment summary}

An overview of the multifocus multiview 3D reconstruction experiment can be summarized as follows:
\begin{itemize}
  \item Capture a set of multifocus single-view images of a calibration dot target facing perpendicular to the optical axis of the lens.
  \item Calibrate camera by detecting dot and computing homography matrices for each image relative to the reference image predefined in the set.
  \item Capture multifocus multiview images of an object of interest without and with back light. 
  \item Stack/blend the multifocus images using the guided filtering algorithm. Apply separately for images without and with back light source.
  \item Threshold the stacked image with back light to create a background mask and add this as a transparent channel of the stacked image without back light.
  \item Obtain the 3D model by feeding the cropped images into the open-source 3D reconstruction software Meshroom \cite{Jancosek2011, Moulon2012, Meshroom}. The 3D reconstruction process includes following steps:
  \begin{itemize}
    \item Camera intrinsic estimation
    \item Feature extraction using SIFT \cite{Lowe04distinctiveimage}
    \item Image matching and feature matching from SIFT descriptors
    \item Structure from motion to generate dense scene and camera poses
    \item Depth map computation
    \item Meshing and Texturing to produce the 3D model
  \end{itemize}
\end{itemize}

\subsection{Camera setup}

Figure \ref{fig:new_scanner_setup} demonstrates the lens and pan-tilt setup for the multifocus multiview imaging experiment. The macro lens (Canon MP-E65mm f/2.8 1-5x) in the setup is connected to a fixed upper-frame, and the camera body (Canon 5DS) is mounted on a macro rail, which is fixed to the lower frame (right of Figure \ref{fig:new_scanner_setup}). Therefore, the lens remains stationary during capturing multifocus images, while the camera body is moving. An expandable rubber duct connects the lens and the camera body to protect the camera sensors from ambient illumination or dust. The target specimen is attached to a pan-tilt turntable (Stackshot 3X) for multiview capturing. For comparison, another experiment setup with the conventional moving lens is also tested (left of Figure \ref{fig:new_scanner_setup} where the macro lens is installed on the camera body, and thus it moves with the camera during capturing). 

In the experiment, a dry flower was chosen as a scanning target. The two experimental setups were performed on the specimen to capture multifocus multiview images and to reconstruct 3D models. Some parts of the flower unfortunately broke off during scanning, so it looks slightly different between the two image acquisitions with different lens setups. 

\subsection{Results and discussion}

\begin{figure}[htbp]
\centerline{\includegraphics[width=\columnwidth]{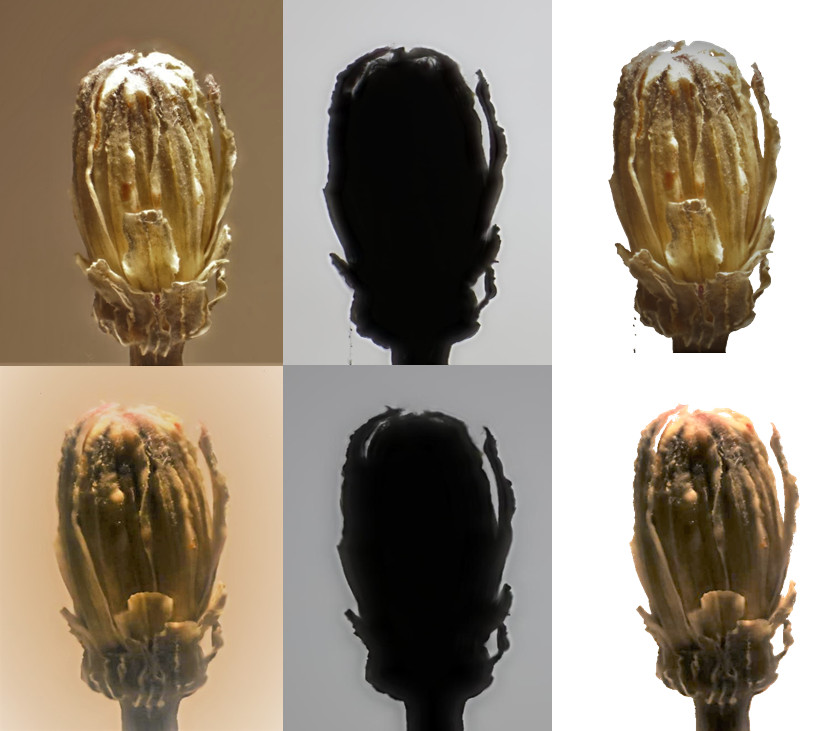}}
\caption{Examples of in-focus images from fixed lens (top) and moving lens (bottom). Left to right: blended in-focus image, in-focus mask and in-focus image with background mask as transparent channel. The in-focus image by fixed lens is sharper than that by moving lens. All images are cropped to show mostly the flower.
}
\label{fig:blending_result}
\end{figure}

To reconstruct the 3D model of the target specimen, image regions belong to backgrounds need to be removed. Therefore, we capture and extract the silhouettes of the specimen by turning on a back-light source. As illustrated in the previous section, the silhouette can be converted into a binary mask by simple thresholding and by adding it to the blended image as a transparency channel. Figure \ref{fig:blending_result} shows an example of in-focus foreground images (left), in-focus back light images (middle) and the final images with background masks (right), for the fixed lens (top) and moving lens (bottom) setups.
From the blending results, it can be seen that the fixed lens setup produces in-focus image with the better surface texture of the flower. 

\begin{figure}[htbp]
\centerline{\includegraphics[width=\columnwidth]{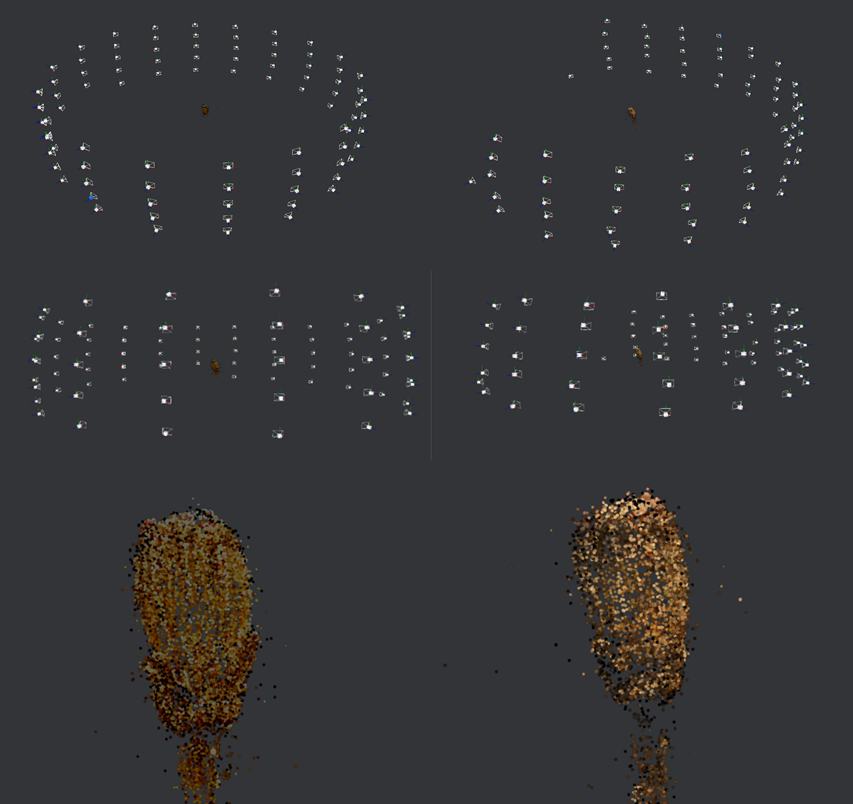}}
\caption{A comparison between structure-from-motion reconstruction from images captured using the proposed Fixed-Lens Multifocus Capture (left) and moving-lens multifocus capture (right). Top row and middle row: camera poses from the reconstruction. Bottom row: sparse point cloud from reconstruction. More uniform camera poses and denser point cloud from fixed lens indicate better image quality than that of moving lens.}
\label{fig:camera poses}
\end{figure}

The blended images (with transparent backgrounds) are then fed into 3D reconstruction software, and the dense clouds and camera poses are generated as shown in Figure \ref{fig:camera poses}. Figure \ref{fig:camera poses} provides a qualitative comparison of the dense clouds and camera poses obtained from the blended images. Although there is a difference in light intensity in input images between two cases, the visual accuracy of the flow model and camera poses using fixed lens is significantly better than that of moving lens. From the dense clouds, the model generated from the moving lens is sparser, and it implies that the conventional moving lens camera setup recovers fewer details of the surface geometry than that of our fixed lens setup. From the estimated camera poses, we can see that the poses reconstructed by the moving lens setup are not well-aligned and missing (only 82 out of 100 camera positions detected). On the contrary, the estimated camera poses generated by the fixed lens can reproduce completed camera trajectory (all 92 camera positions detected), and the recovered camera positions are also well-aligned as expected on a spherical surface.

For a quantitative comparison, we use the estimated camera positions to compute the error in radial position and pan-tilt rotation. Due to pan tilt scanning motion, the camera positions are expected to lie on a spherical surface. A sphere-fit is performed on the estimated camera positions, and the radial distances to the sphere center and pan-tilt angles are obtained. Note that only detected poses are included and carefully processed to avoid introducing artificial outliers. As shown in Table \ref{table:12}, the mean and standard deviation of the radial distances and angles of the poses are provided for both the moving lens and fixed lens. The results show that the estimated camera positions of fixed lens are much closer to the expected values, with estimated panning motion 3.7-times less noisy, than the those of the moving lens.



\begin{table}
\begin{tabu} to \columnwidth { | X[c] | X[c] | X[c] | X[c] |}
 \hline
  & Mean $\pm$ STD of radial distance & Mean $\pm$ STD of pan step & Mean $\pm$ STD of tilt step\\ 
 \hline
 \textbf{Moving lens}: pan step = 18\degree, tilt step = 6\degree & 144.270mm $\pm$ 0.266mm & 18.066\degree$\pm$1.060\degree & 5.708\degree$\pm$0.398\degree \\ 
 \hline
 \textbf{Fixed lens}:  pan step = 20\degree, tilt step = 6\degree & 141.042mm $\pm$ \textbf{0.112mm} & \textbf{20.017}\degree$\pm$\textbf{0.290}\degree  & \textbf{6.273}\degree$\pm$\textbf{0.313}\degree\\ 
 \hline
\end{tabu}
\caption{A comparison between expected and estimated camera centres}
\label{table:12}
\end{table}

\begin{figure}[htbp]
\centerline{\includegraphics[width=\columnwidth]{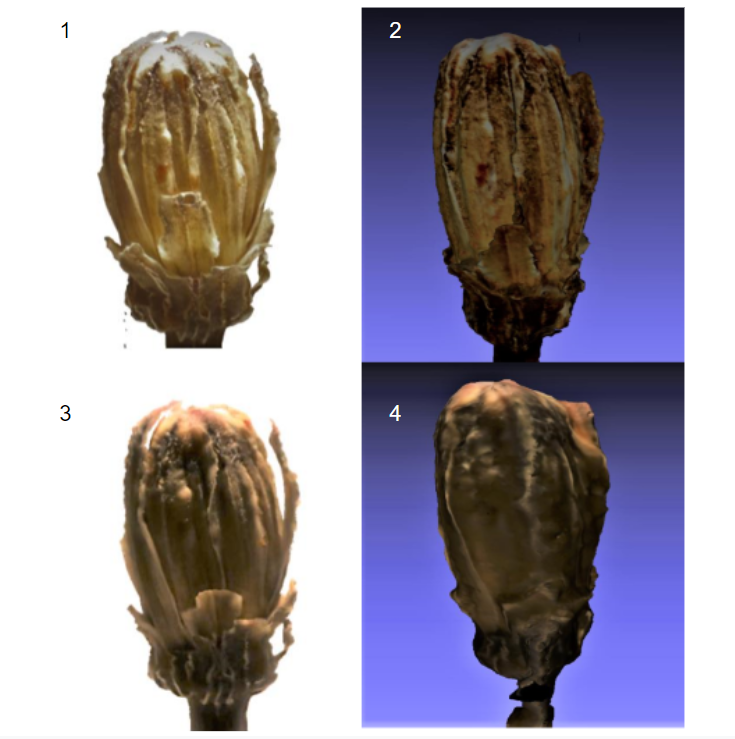}}
\caption{A qualitative comparison between reconstructed 3D models, obtained by fixed lens and moving lens. Top (1 and 2): blended image and reconstructed 3D model obtained by fixed lens. Bottom (3 and 4): blended image and reconstructed 3D model obtained by moving lens.}
\label{fig:mesh}
\end{figure}

Figure \ref{fig:mesh} shows a pair of in-focus image and snapshots of reconstructed models from a similar angle. From image 2 and 4 in Figure \ref{fig:mesh}, it can be seen that the model obtained by the fixed lens has a significantly better visual accuracy on the reconstructed surface geometry of the target specimen. Finer details and textures of the specimen's surface are preserved in the model from the fixed lens than those from moving lens. This matches with the 3D reconstruction result shown in Figure \ref{fig:camera poses}, where the fixed lens setup produces a more completed and more accurate estimation of camera positions and dense clouds. However, deviations between the source image and reconstructed model still exist. For example, either of the models cannot reconstruct the gaps between leaves in the source images. The deviations might be caused by limitations of the 3D reconstruction software, as micro-structures (such as transparent wings of insects or the tiny gaps in the images) are usually very difficult to be recovered. This defective reconstruction indicates the direction of future research, which could focus on recovering the micro-structures of specimens by testing different 3D reconstruction technique or software. 

\section{Conclusion}
In this paper, we proposed an image-based 3D reconstruction setup using the proposed Fixed-Lens Multifocus Capture and the homography-based scale-shift calibration for accurate 3D reconstruction of small-scale objects. Currently, image-based 3D reconstruction devices using lens moving with the camera to capture multifocus images suffer from perspective inconsistency that reduces the accuracy of image stacking and 3D reconstruction. With the proposed Fixed-Lens Multifocus Capture setup, the lens remains stationary while the camera and image sensor move during multifocus image capturing. The scale-shift image calibration via homography estimation is performed to account for the change of relative scale and in-plane shift between images captured at different depths. Multifocus images corrected for scale-shift change are stacked/blended to create an in-focus image that is consistent with the perspective image formation. 

Initial multifocus multiview 3D reconstruction experiments were performed to demonstrate the effectiveness of our proposed fixed lens setup as compared with conventional moving lens setup. The reconstruction results show that the proposed Fixed-Lens Multifocus Capture generated a more accurate 3D model and more accurate camera poses (at least three times less noisy) than those by the conventional moving lens. This can be explained by the fact that the perspective inconsistency in in-focus images captured using moving lens violates pinhole camera model assumption of 3D reconstruction and this leads to reconstruction errors and quite a few camera positions are missing. By using proposed Fixed-Lens Multifocus Capture to generate in-focus images that preserve perspective image formation, the multiview 3D reconstruction works as expected, detected all camera positions and created a more accurate 3D model.




\Urlmuskip=0mu plus 1mu\relax
\bibliographystyle{plain}
\bibliography{dicta2019_3Dmultifocus_references}

\end{document}